\documentclass{article}

\usepackage{PRIMEarxiv}

\usepackage[utf8]{inputenc} 
\usepackage[T1]{fontenc}    
\usepackage{hyperref}       
\usepackage{url}            
\usepackage{booktabs}       
\usepackage{amsfonts}       
\usepackage{nicefrac}       
\usepackage{microtype}      
\usepackage{lipsum}
\usepackage{fancyhdr}       
\usepackage{graphicx}       
\graphicspath{{graphics/}}     

\usepackage{makecell}
\usepackage{caption}
\usepackage[super]{nth}
\usepackage{lmodern}
\usepackage{algorithmic}
\usepackage{algorithm}
\usepackage{amssymb} 
\usepackage{aliascnt}
\usepackage{mathtools}
\usepackage{changebar}
\usepackage{ifpdf}

\hypersetup{
hypertexnames=false, 
bookmarksnumbered=true,
colorlinks=false,
pdfborder={0 0 0},
anchorcolor=black,
linkcolor=black,
citecolor=black,
filecolor=black,
urlcolor=black,
menucolor=black
}

\usepackage[compact]{titlesec}
\usepackage[title]{appendix}

\usepackage{etoolbox}
\makeatletter
\patchcmd{\chapter}{\if@openright\cleardoublepage\else\clearpage\fi}{}{}{}
\makeatother

\title{Automated Repair of Neural Networks}

\author{
  Dor Cohen, Ofer Stricman \\
  Department of Information Sytems\\
  Technion---Israel Institute of Technology\\
  Technion City, Haifa 32000, \underline{Israel}\\
  \texttt{dor.coh@gmail.com}, \texttt{ofers@technion.ac.il}
}

\begin{document}
\maketitle

\begin{abstract}
Over the last decade, Neural Networks (NNs) have been widely used in numerous applications including safety-critical ones such as autonomous systems. Despite their emerging adoption, it is well known that NNs are susceptible to Adversarial Attacks. Hence, it is highly important to provide guarantees that such systems work correctly. To remedy these issues we introduce a framework for repairing unsafe NNs w.r.t. safety specification, that is by utilizing satisfiability modulo theories (SMT) solvers. Our method is able to search for a new, safe NN representation, by modifying only a few of its weight values. In addition, our technique attempts to maximize the similarity to original network with regard to its decision boundaries. We perform extensive experiments which demonstrate the capability of our proposed framework to yield safe NNs w.r.t. the \textit{Adversarial Robustness} property, with only a mild loss of accuracy (in terms of similarity). Moreover, we compare our method with a naive baseline to empirically prove its effectiveness. To conclude, we provide an algorithm to automatically repair NNs given safety properties, and suggest a few heuristics to improve its computational performance. Currently, by following this approach we are capable of producing small-sized (i.e., with up to few hundreds of parameters) correct NNs, composed of the piecewise linear ReLU activation function. Nevertheless, our framework is general in the sense that it can synthesize NNs w.r.t. any decidable fragment of first-order logic specification.
\end{abstract}


\section{Introduction}
\label{chap:intro}

In the last decade artificial \textit{deep neural networks} (NNs) \cite{goodfellow2016deep} have played a significant role in advancing research in diverse fields such as computer vision \cite{krizhevsky2012imagenet} and natural language processing \cite{vaswani2017attention}, surpassing the human level in a variety of tasks. However, it is well known that these complex models also carry major threats which restrict their integration in safety-critical systems. Not long ago, \cite{szegedy2013intriguing} observed that NN models which excel in image recognition tasks can be fooled by injecting small but calculated perturbations to their inputs. At the same time, \cite{biggio2013evasion} showed how an NN-based controller for malware detection could be evaded by a similar type of attack. Since then, enormous amounts of methods were developed through what can be described as an ongoing arms race between attackers and defenders - as demonstrated in \cite{athalye2018obfuscated} who stressed how the proposed defenses could be easily broken, while applying attacks which were not considered by the designers. 

These challenges have paved the way for the development of verification methods which can provide formal guarantees on the behavior of NNs \cite{katz2017reluplex}, \cite{wang2018formal}, \cite{ehlers2017formal}, although such methods are still limited in their ability to verify modern networks properties, for a few reasons. First, NNs model highly nonlinear operations as logistic or hyperbolic tangent functions, and as such, most NN verification methods (according to a comprehensive survey \cite{liu2019algorithms}) are specialized in network architectures composed of piecewise linear functions. Second, NN architectures are very often tremendously large, and may contain millions of learned parameters \cite{devlin2018bert}, casting the problem intractable unless using approximating methods. Recently, \cite{katz2017reluplex} proposed a method for exact verification of NNs properties which scales to a few thousand parameters. Once applied, such verification methods are able to prove if a certain property holds for all inputs within a given range (e.g., $x \in [0, 0.5] \implies f(x) \leq 0.3$), otherwise they provide a counterexample. Despite their ability to generate many property violating inputs which can hypothetically be used to refine the network safety, verification methods were not yet employed to supply information on how to repair the model (i.e., change its parameters) in order for it to meet a required property.

The robustness of machine learning (ML) methods is determined by following the statistical learning theory \cite{valiant1984theory}. It provides a probabilistic guarantee that the learned model test error rate is bounded by some threshold. More concretely, the algorithms used to calibrate ML models are fed with a finite set of input-output examples, also called \textit{training set}, which is assumed to represent the probability distribution of the true (but unobserved) data-generating model. To evaluate the performance of a trained model, a small randomly-sampled subset of the data is kept aside and used to measure the (empirical) \textit{test set} error rate of the model. Thus, trained models with low test error rates are expected to generalize, namely, evaluate previously unseen data (assumed to be generated from the true model), correctly. However, relying on such guarantees when considering safety applications that require specific properties to hold is inherently not sufficient. Frequently, when a certain NN does not meet the required specification, for instance, \textit{adversarial robustness} (i.e. a property which requires the model decision to be stable within a small neighborhood of an input sample), practitioners re-train the learned models using misclassified inputs as new training data samples, expecting the new model to better generalize \cite{goodfellow2014explaining},\cite{madry2017towards}. This technique is also known as \textit{adversarial training}, forming the basis of many suggested defenses. \\

In this work, we aim to mitigate the challenging process of refining NNs to maintain safety properties. Instead of iterating between verification and retraining methods until the model is safe, we suggest to directly search for the model parameters by exploiting formal methods. We propose a framework to automatically and provably repair a previously trained (unsafe) NN w.r.t specification, while keeping the repaired function representation similar as possible to the original one. First, we train our model on a designated task using traditional algorithms, and then we utilize \textit{satisfiability modulo theories} (SMT) \cite{shostak1979practical} to formulate our problem and search for the correct parameters for which the model will satisfy the required specification. We apply our framework to repair the \textit{adversarial robustness} property of small-sized trained NNs, and demonstrate its capability to produce secured NN-based controllers. In contrast to \textit{adversarial training} which provides bounds on the test error rate, our framework yields an exact safe network representation and is general in the sense that it can enforce any decidable fragment of first-order logic specification.

\section{Preliminaries}
\label{chap:prelims}

\subsection{Verification of Neural Networks}

Verification methods allow to prove whether functions fulfill a certain specification - i.e., relation between a function's input and output. Such a property can be formulated in first-order logic as follows, for a given parameter set $w$: 

\begin{equation}
\label{eq:equation1}
\forall x \;\; C_{in}(x) \; \implies \; C_{out}(f_w, x)
\end{equation}

where $C_{in}$, $C_{out}$ denote input and output constraint functions respectively, $x$ is an input to the system, and $f_w$ is a parameterized function which represents the system output (i.e., $y=f_w(x)$). Various search-based methods for NN verification such as \cite{katz2017reluplex}, exploit the fact that $\forall x. F$ is equal to $\lnot \exists x . \lnot F$, where $F$ denotes a desired property \footnote{ For example, consider the following encoding:  $\forall x \in [l,u]   \;.\;\; f_w(x) \leq th $ , its corresponding negation is  $ \exists x \in [l,u] \;.\;\; f_w(x) > th$. This equivalence allows to eliminate the universal quantifier, that is by transforming the formula into $\exists x \; f_w(x) > th \land x >= l \land x <= u $.} (e.g., input-output specification). Under this simplification, if the search returns an assignment \footnote{ In first-order logic a solution to a formula must give an interpretation to unquantified (free) variables. Therefore, to return an assignment for $x$, the formula should not include quantifiers which involve $x$. If it does and $x$ is the only variable, then its solution is simply True or False (e.g., if there exists such $x$ which satisfy certain constraint or not). Throughout our formulations, we keep the quantifiers for the purpose of clarity, even though they are not always required.} which satisfies the formula, then this assignment is a counter-example which proves the specification does not hold . Therefore, to find if property \autoref{eq:equation1} is not safe, we can check the negation of its formula. Consider the case where we have a function $f_w$ which has an unsafe property encoded by \autoref{eq:equation1}, it can be extended to search for a new model $f_{\hat{w}}$ with different parameter set $\hat{w} \neq w$ which may satisfy the required specification (assuming such model exists), that is by adding an existential quantifier: 

\begin{equation}
\label{eq:equation2}
\exists \hat{w} \;\; \forall x \;\; C_{in}(x) \; \implies \; C_{out}(f_{\hat{w}}, x)
\end{equation}

Naturally, solving \autoref{eq:equation2} is harder, since we consider multiple function representations (defined by different parameters) to satisfy the property, while in \autoref{eq:equation1} we only consider a certain constant function. More formally, it was proved in \cite{katz2017reluplex} that verifying the robustness property for NN (composed of only piece-wise linear operations, namely ReLU functions) is NP-Complete. Considering formula \autoref{eq:equation2} however, transforms this problem complexity class into $\Sigma-2$ Complete.

\subsection{SAT and SMT}

\subsubsection{SAT}
Let $x_1, ..., x_n$ be Boolean variables. The SAT problem is defined as deciding whether there exists an assignment which satisfies a formula $\phi$ consisting of Boolean logic constraints on $x_1,...x_n$. We assume that  $\phi$ is given in \textit{conjunctive normal form} (CNF). Then, by employing the DPLL \cite{davis1962machine} algorithm - a backtracking-based search procedure, the solver is able to decide the satisfiability of $\phi$. DPLL works by deciding on the variable assignment in each level of the search tree (i.e., partial assignment to $\phi$). In case of a conflict, it applies resolution and updates $\phi$, to ensure the algorithm will not reach again this conflict, and then it backtracks to the last partial assignment. The algorithm returns SAT if a satisfying assignment was found (full assignment is represented by a leaf in the search tree), or UNSAT if all full assignments are conflicted. For example, consider $x_1, x_2, x_3$ as our variables and $ \phi = (x_1 \lor x_2) \land (\lnot x_1 \lor x_3)$. At first, the algorithm may decide $x_1=True$, then $\phi = (True \lor x_2) \land (False \lor x_3) = x_3$, at this point, it decides $x_3 = True$ and $x_2$ could take any value. In this case, the algorithm will return SAT along with an assignment (e.g., $x_1=True, x_2=True, x_3=True$). 

\subsubsection{SMT}
is a method for proving the satisfiability of decidable fragments of first-order formulas under a decidable theory (i.e., for which a method to derive the correct answer exists). SMT extends the boolean satisfiability (SAT) problem by supporting more theories (e.g., theory of linear real arithmetic, theory of bit-vector arithmetic, etc.). Fundamentally, in order to solve other theories, SMT solvers (i.e. programs which solve SMT) generate an abstraction by replacing each constraint with a boolean variable. Then, supported by a corresponding theory solver, the method operates similarly to DPLL, deciding on assignments for variables and refining the abstraction (using the relevant theory solver) through the search process. As an example, consider the real variables $x$,$y$ in the following first-order formula: $\phi = (y > 3x) \land ((x >= 1.5) \lor (y >= 1.5))$. We can replace each linear equation with a boolean variable: $\phi = A \land (B \lor C)$. Then, assuming the solver decided $B=True$, it implies $\phi = A \land B$, the theory solver is then executed to check the feasibility of $\{ y > 3x, x >= 1.5 \}$, which after simplification is equal to $y > 4.5$. In this example, the solver returns SAT, possibly with the following assignment: $x=1.5, y=5$.

SMT solvers allow us to solve formulas similar to \autoref{eq:equation1}, \autoref{eq:equation2}. However, adding quantifiers to the formula generally makes its theory harder for solving, where in some cases it could even become undecidable (e.g., in the case of quantified non-linear integer arithmetic). In this scenario, we do not have a method to decide the formula satisfiability. In our setting, we will generally be using the theory of \textit{non-linear real arithmetic} (NRA), since in formulation \autoref{eq:equation2} we expect to have constraints including multiplications of variables. \textit{Cylindrical algebraic decomposition} (CAD), an algorithm which was originally introduced by \cite{collins1975quantifier} to solve \textit{quantifier elimination} (QE) and is employed at the core of Z3 \cite{de2008z3} non-linear SMT solver, has double-exponential complexity in the number of formula variables \cite{davenport1988real}. One can find more details on QE and CAD in \cite{cadtut18}. In our trials, we also experimented with QE methods for reducing the theory to \textit{Quantifier-Free NRA} (QF\_NRA), expecting to earn more efficiency; although, unfortunately, the methods implementing QE \cite{collins1991partial},\cite{dolzmann1997redlog} were not scalable enough even in the case of our small models. Lastly, we also considered experimenting with Binarized NN \cite{courbariaux2016binarized}, for the same reasons.

\subsection{Neural Networks}
A neural network in its base form is a mapping $f_w: X \rightarrow Y$ where $X \in \mathbb{R}^{n}$, $Y \in \mathbb{R}^{m}$,  $n,m \in \mathbb{N}$ and $w$ denotes the set of function parameters. This mapping usually forms a composition of $k$ functions: $f_w(x) = f^k(...(f^2(f^1(x))))$, where $k$ is also referred to as the number of layers. Typically, each function $f^i$ performs a linear transformation followed by non-linear one: $f^i(x) \equiv g_i(W_ix+b_i)$, where $g_i$ denotes the non-linear operation, also known as \textit{activation function}, $W_i$, $b_i$ are matrix and vector of parameters, and their shape indicate the number of neurons for specific layer $i$. In our setting, we consider the common ReLU activation function which is piece-wise linear: $g(x) = max(0, x)$. For example, a two-layer NN with $d$ ReLU neurons, an input $x\in \mathbb{R}^n$ and output $y \in \mathbb{R}^m$ will have the following form \footnote{For the sake of simplicity, in our framework we exclude the non-linear operation from the last layer, instead we set it to the identity function: $g(x) \equiv x$. Often, this non-linear operation is normalized exponential function (softmax), for example it transforms the vector $\{2,3,5\}$ into $\{0.042,0.114,0.844\}$. Then omitting this operation will require no changes in the encoding when considering robustness properties.}: $f_w(x)=W_2*max(0,W_1x+b_1)+b_2$, where $W_1 \in \mathbb{R}^{d \times n}$, $b_1 \in \mathbb{R}^d$, $W_2 \in \mathbb{R}^{m \times d}$, $b_2 \in \mathbb{R}^{m}$.

\subsubsection{Adversarial robustness of NNs}
Adversarial robustness defines invariance to noise in a $\delta$ neighborhood of  $x_0$ as follows:

\begin{equation}
\label{eq:equation3}
\forall x . \; \| x-x_0 \| \leq \; \delta \; \implies \; D(f_w(x)) = D(f_w(x_0)),
\end{equation}
where D denotes the decision function, a mapping required in classifiers (i.e., ML models which map inputs into discrete categories), usually $D(f(x)) = \underset{j}{\operatorname{argmax}}  (f(x))$ where $j$ denotes the model predicted category.

\section{Related Work}
\label{chap:related}

We first note that up to the starting point of this research work, we were not aware of previous works which attempt to solve the general problem of synthesizing a provably correct NN. However, throughout the work, a few related works were published. 

Thus, in the following we introduce benchmarks for evaluating verification methods for NNs properties, we review a few verification methods and describe how our framework complements them. Then, we briefly survey the development of adversarial techniques in the ML community and their relation to our framework. Lastly, we describe recent novel methods which are intended to produce provably correct NNs.

\subsection{Formal Verification of Neural Networks}

\begin{figure}[htb]
	\centering
	\includegraphics[scale=0.35]{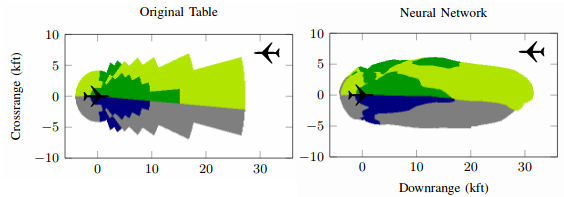}
	\caption{Taken from the \cite{julian2016policy} - depicted at left are the decision boundaries derived from the original \textit{ACAS Xu} hash table. At the right figure, we can observe the trained NN boundaries. Each color depicts a different action that the aircraft should carry, with respect to the intruder position. It can be seen how the NN imposes unsafe regions, in contrast to the original table.}
\end{figure}\label{fig:acas}

In \cite{julian2016policy} it was proposed to compress hash tables which store large state spaces into NNs, effectively reducing their storage size by a factor of 1000 (i.e., from 2GB to 2MB). They demonstrate this reduction on \textit{ACAS Xu} - an anti-collision controller for unmanned aircrafts. More specifically, it takes as input sensor data which indicates speed and direction of an aircraft and nearby intruders, and then it advises the aircraft which action it should follow in order to avoid collision (e.g., intruder approaches from right $\implies$ turn hard left). Consequently, there was a need to verify that these models still satisfy their safety properties after the reduction, as any violation could lead to an unsafe system, as can be seen in Fig \ref{fig:acas}. \cite{katz2017reluplex} proposed Reluplex, a sound and complete method for verifying properties of NNs with ReLU activation functions. Their method searches for counterexamples in a depth-first manner. It builds upon a specialized version of the Simplex algorithm to set an assignment iteratively, along with an SMT solver for lazily branching on neurons which their state violates the property too often in the search,  performing exhaustive enumeration on all possible combinations of variables states\footnote{ReLU neurons have two states. They are active when $g(x)=x$, and non-active when $g(x)=0$. } in the worst case. Reluplex was evaluated on various properties that an \textit{ACAS Xu} NN-based controller should satisfy, and proved to be scalable for this specific task (the model had $\approx$ 13,000 parameters). Recently, a successor of Reluplex was published under the name Marabou \cite{katz2019marabou} which mainly provides more efficient search heuristics. Another method called ReluVal \cite{wang2018formal} offers a sound but incomplete procedure to prove properties by symbolic interval propagation. Given the desired input and output ranges, it computes upper and lower bounds for each neuron, propagating their (bounded) values layer by layer to finally produce an overapproximation of the network output values. The main distinction between these methods, including many others surveyed in \cite{liu2019algorithms}, is that they consider only constant NNs (i.e., with fixed parameters), while in our framework we must solve quantified formulas to declare the parameters as free variables. Our framework complements such methods, as we would want to check if the required properties hold before executing an automated repair. 

\begin{figure}[htb]
    \centering
	\includegraphics[scale=0.2]{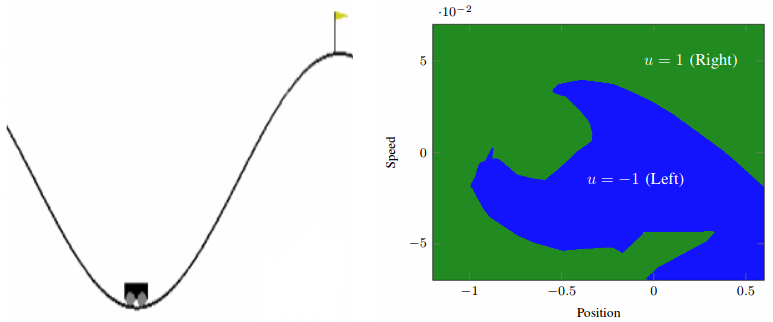}
	\caption{Depicted at left is an illustration of the mountain car problem. In this task, a vehicle gets an initial position and velocity, and then in each step, it needs to take the best action (steer right, steer left, or do nothing) which would result in reaching the flag quickly as possible. At right, we can see the decision boundaries (i.e. speed as a function of position) of the NN-based controller, aimed at solving this task \cite{julian2019reachability}.}
	\label{fig:mcar}
\end{figure}

Another recent method \cite{julian2019reachability}, which over-approximates reachable sets of the NN output (similar to ReluVal), verifies a NN controller which was trained to solve the \textit{mountain car problem}, more detailed in Figure \ref{fig:mcar}. This controller actions domain is smaller than that of \textit{ACAS Xu}, specifically, it contains 3 possible actions, additionally, its input domain is $\mathbb{R}^2$. The authors use a NN consisting of 5 layers with 30 ReLU neurons each, to model this controller behavior.

\subsection{Adversarial Examples}

Adversarial attacks and defenses for NNs are thoroughly studied in the literature. Since  \cite{szegedy2013intriguing} and \cite{biggio2013evasion} introduced the first methods, \cite{goodfellow2014explaining} proposed \textit{fast gradient sign method} (FGSM) attack, which was efficiently integrated in an \textit{adversarial training} setting with gradient descent based algorithms, producing more empirically robust models. Then, \cite{madry2017towards} suggested training with \textit{projected gradient descent} adversary, which is considered as a stronger attack since it allows to search for violating example in multiple gradient steps, in contrast to FGSM  which consider only one step look-ahead search. Even though these defense methods are all empirically robust, providing lower bounds on the (adversarial) test set error rate \footnote{An adversarial, or robust test set, is a set of ordinary input samples which were manipulated using adversarial attacks, to intentionally cause misclassification. These sets are often used as benchmarks for evaluating adversarial defenses.}. Lately, few works also proposed \textit{certified defenses}, given a family of attacks, they ensure no perturbation exists within an $L_p$ norm ball of radius $\epsilon$ which causes misclassification. In our work, we aim to bypass the stage of \textit{adversarial training} by directly searching for model parameters which satisfy certain specification. Also, in contrast to methods such as \cite{wong2017provable},\cite{weng2018towards} which provide \textit{certified defenses} on norm-bounded attacks (i.e., perturbed input samples which are bounded within their $L_{\infty}$ norm), our framework allows to enforce any specification which can be formulated in a decidable fragment of first-order logic.

\begin{figure}[htb]
	\centering
	\includegraphics[scale=0.2]{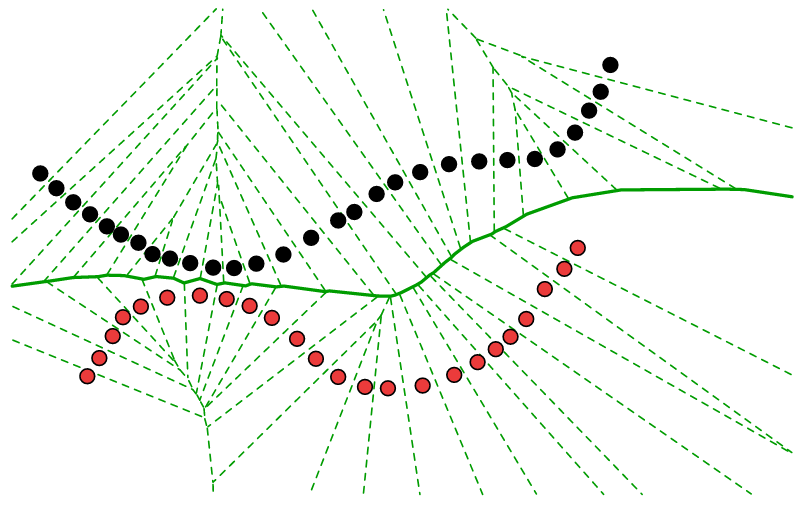}
	\caption{Borrowed from \cite{khoury2019adversarial}, this figure depicts the Voronoi cells of samples from two different classes, distinguished by red and black. Each sample lies within its own Voronoi cell depicted in green. The bold green line indicates the decision boundary, determined by adjacent cells of different classes. Considering Voronoi cells ensures the boundaries to be far as possible from any training sample, intuitively leading to more robust models.}
	\label{fig:voronoi}
\end{figure}

A recent defense that employs \textit{adversarial training}, suggests to explicitly push the decision boundary to be far from the training data manifold, claiming to result in more robust models \cite{khoury2019adversarial}. Specifically, they replace the $L_p$ norm ball as shown in \eqref{eq:equation3} with Voronoi cell at each sample $x$. The Voronoi cell of $x$ is the set of all points in $\mathbb{R}^n$ that are closer to $x$ than any other sample in $X$ (i.e., our training set). An example for such cells for $x \in \mathbb{R}^2$ can be seen in Figure \ref{fig:voronoi}. The main advantage in this setting is that, by definition, Voronoi cells do not intersect with each other, in contrast to $L_p$ norm balls which may intersect for certain values of $\epsilon$. 

\subsection{Provably Correct Neural Networks}
Lately, few works which attempt to solve the general problem of producing correct neural networks were published. In \cite{lin2020art}, the authors suggest producing correct NNs by construction (i.e., via the training phase). In their framework, the required safety constraints were encoded as differentiable functions, allowing the constraints to be incorporated into the training objective function. More concretely, their objective is composed of both accuracy and safety terms, although it also imposes a trade-off between the two goals. Since both terms are differentiable, traditional gradient descent solvers can be utilized to optimize the network parameters. During the training phase, once the safety term of the objective function reached zero, it is guaranteed that the property holds. However, generally, no convergence is guaranteed when optimized using gradient-based methods. And yet, \cite{lin2020art} demonstrated how their framework can produce safe \textit{ACAS Xu} networks, with a mild loss of accuracy. In another work, \cite{mell2020safe} propose to train multiple learners, where each one is accountable for a separate part of the decision boundary. Then, when combined together using learned weights, these predictors assemble a safe model. Both works employ gradient methods in their frameworks. Our work, however, utilizes SMT solvers and focuses on finding a safe network that is the most similar to the original one.

Another line of work applies SMT solvers to produce correct NNs, as in this work. \cite{sotoudeh2019correcting} introduce a framework for patching NNs by modifying a small subset of the NN weights. They suggest to manually select 2-dimensional areas of the NN decision boundaries (e.g., a fixed polygon) while other dimensions are kept fixed, then they compose a specification following these selections. The authors reduce the nonlinear constraints into linear ones by using approximation methods to represent the original network, this allows them to employ linear programming (LP) solvers. Their results demonstrate that applying these patches, in addition to providing safety, might also cause the network to generalize to the training data. In \cite{papusha2020incorrect}, the authors propose bias patching - i.e., modifying only values of bias weights. Specifically, they search for a value which can be added to each bias parameter in the network, such that the safety constraints will hold. To perform such a search, they use a quantized NN (i.e., a compact representation of the original NN with decreased precision). Our work, in contrast, attempts to automatically repair the original NN w.r.t. any decidable fragment of first-order logic specification, while theoretically being able to free any number of weights.

Lastly, \cite{leino2021self} introduce a framework to produce safe NNs without actually modifying the original network, rather, they suggest appending another layer to the network. This extra layer is accountable for enforcing safety specifications by modifying outputs that violate the desired property. It allows them to decouple safety from the process of training (or repairing) the network.

\section{Provably Repair a Neural Network}
\label{chap:problem}

\subsection{Problem Definition}\label{chap:problem-def}

\textbf{Problem definition:} Consider a NN model denoted by $f_{w_A}(x)$ (model $A$). Assume that $f_{w_A}(x)$ is not safe w.r.t. a required specification $\varphi$. In the task of provably repairing a NN, we search for a new model $f_{w_B}(x)$ (model $B$) which satisfies $\varphi$ and its decision boundaries are similar as possible to model $A$. 

We assume that the learned model $A$ already represents the desired function approximately, meaning it may violate $\varphi$ but satisfy other required properties. Additionally, for computational reasons, we repair a few properties at a time. Therefore, we prefer to initialize most of model $B$ parameters values with these of model $A$, and have very few free parameters in our formula. For this purpose, we decompose model $B$'s set of weights denoted as $w_B$ into two components $w_B = w_{B-const} \cup w_{B-free}$, where $w_{B-const}$ refers to the set of constant parameters (initialized using model $A$), and $w_{B-free}$ is the set of parameters we modify.

\subsubsection{Use case: Adversarial Robustness property}

Without loss of generality, let $\varphi$ be the \textit{adversarial robustness} property with respect to some sample point $x_0$ and a $\delta$-neighborhood of it, and assume that model $A$ does not satisfy $\varphi$. To generate a repaired network $f_{w_B}$  which satisfies $\varphi$, we search for a subset of weights $w_{B_{free}} \subseteq w_B $, and corresponding values, such that

\begin{equation}
\label{eq:equation4}
\forall x . \;\;\; \| x-x_0 \| \leq \; \delta \; \implies \; D(f_{w_B}(x)) = GT(x_0),
\end{equation}

holds, where $f_{w_B}$ is a NN in which the weights in $w_{B_{free}}$ are set to those values, and the rest are set to the values of the corresponding weights in the original NN.

In this formulation, we require model $B$ to classify $x_0$ according to the ground truth ($GT$). As noted earlier, we set model $B$ parameters with their original (model $A$'s) values, and then we choose to free only few parameters (i.e. $w_{B_{free}}$), hence $f_{w_B}$ share most of $f_{w_A}$ weights. Moreover, to direct model $B$ to be similar as possible to model $A$, there are a few tactics that we considered. We can add more constraints to our formulation such as, e.g., force model decision at certain points: $x=x_0 \implies D(f_w(x)) = GT(x_0)$. Another direction is to explicitly encode the original decision boundaries into discrete cells and then guide the search towards minimal violations. We give more details about it later in \autoref{main:exp-setup}. Additionally, a complete formula encoding for enforcing a specification on a simple NN is detailed in \autoref{appendix:basic_nn}. 

Lastly, note that \autoref{eq:equation4} does not describe an optimization problem, rather, it provides a logical formula which asks whether there exists a subset of weights that can be replaced to assemble a safe network. Assuming there are multiple subsets that satisfy the safety properties, then in order to find the most accurate solution, we suggest optimizing over this formulation. Therefore, we later propose a method (\autoref{alg:first}) for finding a repaired model whose decision boundaries are similar as possible to the original NN. Such optimization is essential since $w_{B_{free}}$ can represent any valid combination of free weights.

\subsection{Initial Experiments}

\subsubsection{Provably Repair a NN}\label{main:initial-exp}

Consider the $XOR_{mini}$ dataset. It has input $x \in \mathbb{R}^2$ and output $y \in \{0,1 \}$. Specifically its data points and their labels correspond to the following mapping: $T_{XOR_{mini}} = \{(10, 10): 0, (-10, -10): 0, (-10, 10): 1, (10, -10): 1 \}$. Let model $A$ denote two-layer NN with 8 neurons that was pre-trained with common gradient descent search algorithm on the $XOR_{mini}$ dataset. We experimented with solving \eqref{eq:equation4} in order to find a repaired model $B$ which satisfies the robustness property w.r.t. points in $T_{XOR_{mini}}$.

\begin{figure}
	\centering
	\includegraphics[scale=0.275]{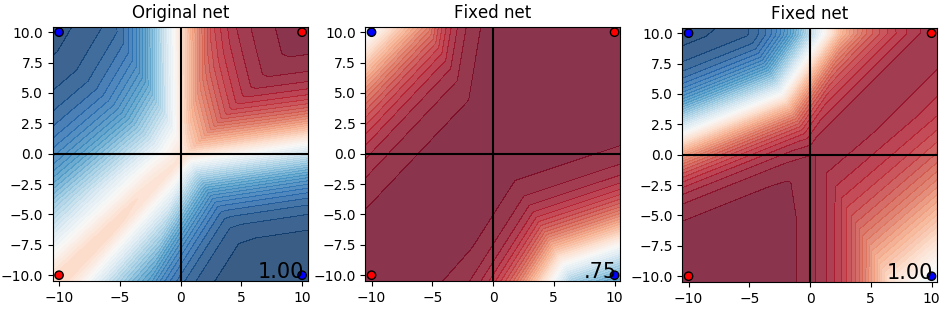}
	\caption{Depicted at left is model $A$ decision boundaries. The middle and right figures depict two repaired models w.r.t. point $(-10, -10)$. In contrast to the middle figure, the right one attempts to retain part of the original decision boundary. The training set accuracy metric is depicted at the bottom right corner of each figure.}
	\label{fig:initial-repair}
\end{figure}

In particular, we chose to repair the model w.r.t. the following sample $(x_1,x_2) = (-10,-10)$, to satisfy the robustness property in its $\delta=8$ neighborhood, defined by $L_1$ norm. In \autoref{fig:initial-repair}, we can see how the original model does not satisfy the property, in contrast to the two repaired models, although the repaired boundaries have changed significantly, obviously as we have free parameters in our formula. To avoid this unwanted behavior, we may take different steps to retain parts of the original boundaries. First, we can consider bounding the new parameters in a neighborhood of their original values, intuitively staying close to the first form. However, model $B$ parameters do not necessarily have to be close to the originals. Another direction is to store, in some way, the original decision boundary. Generally, we want to keep the size of the formula as small as possible. Hence the challenge is to store the boundaries in an efficient way. At this point, we found a simple heuristic, by which we add as constraints model $A$ decisions for each of the points in $T_{XOR_{mini}}$, as can be seen in the rightmost figure. Additionally, even if we have an efficient method to store the boundaries as constraints, we must soften them, since we do expect our model representation to slightly change, as it does not meet a certain specification in the first place. One option to tackle this is by casting our problem as an optimization problem. If we wrap the boundary constraints in Boolean indicators, we can then search for a solution that maximizes their sum. In this manner, the repaired model is ensured to be as similar as possible to model $A$. 

The value for $\delta$ was determined by manual experiments: it was the maximal value that is lower than 10, which the solver returned SAT. In addition, we chose to free one weight and one bias parameter in the first neuron of the first layer. Lastly, we can also see from \autoref{fig:initial-repair} how the repair affected the confidence level of our model, which is depicted by the color boldness (i.e., dark color depicts high confidence and vice versa). This violation could also possibly be fixed by assisting with the original network decisions.

\subsubsection*{Technical Details}
Through the initial trials, we use Z3 \cite{de2008z3} version $4.8.12$ as an SMT solver and its Python wrapper for generating the abstractions. Z3's nonlinear theory solver is described in \cite{jovanovic2012solving}. PyTorch \cite{paszke2017automatic} was used for training the NN, where we employ Negative Log-Likelihood (NLL) loss function optimized using Stochastic Gradient Descent (SGD), with a learning rate equal to 0.1, and a total of 10 epochs (i.e., training iterations).

\subsection{Experimental setup}\label{main:exp-setup}
The initial experiments demonstrate how SMT solvers can be utilized to automatically repair unsafe NNs. However, our proposed framework entails a few challenges. Namely, storing the NN decision boundaries, to find a repaired model which is similar as possible to the original, or, another challenge is to ensure that the repaired network confidence level is still reasonable. To try and tackle these problems, our work explores a few directions. To maximize the similarity to the original model, we shall add soft constraints to our framework, allowing us to solve a maximization problem. Then, on top of it, we shall use several heuristics as described below.

To exemplify these points, we perform two main sets of experiments. In the first set, we compare our considered similarity-preserving heuristics. Furthermore, the second set of experiments demonstrate how our method contrasts a naive baseline. Thus, in this section we describe the benchmarks used in these experiments  (i.e., datasets, NN models, and their examined specifications), then we give additional details regarding our considered similarity heuristics.

\subsubsection{NN Architectures, Datasets, \& Specifications}\label{subsection::setup}

\begin{figure}
    \captionsetup{font={footnotesize}}
	\centering
	\textbf{Examined Neural Networks topology illustrations}\par\medskip
      \begin{minipage}[b]{0.4\textwidth}
        \includegraphics[width=\textwidth,scale=0.4]{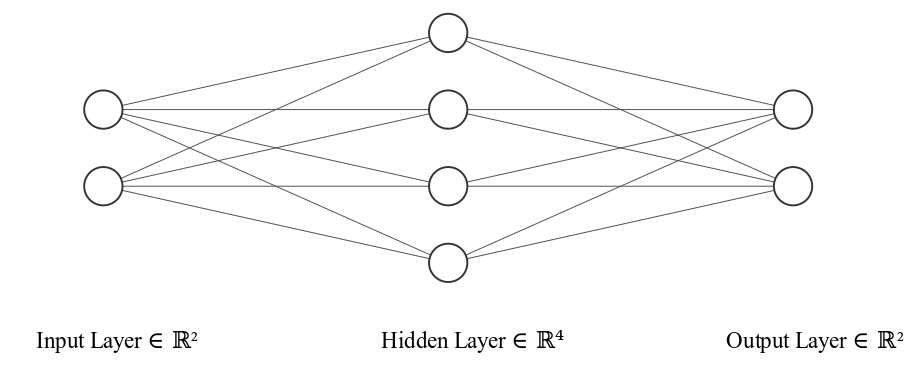}
        \caption{The XOR architecture has one hidden layer with 4 ReLU neurons. Models XOR-A and XOR-B were both originated from this topology.}
        \label{fig:topology-xor}
      \end{minipage}
    \hspace{1 cm}
    \begin{minipage}[b]{0.4\textwidth}
        \centering
        \includegraphics[width=\textwidth,scale=0.55]{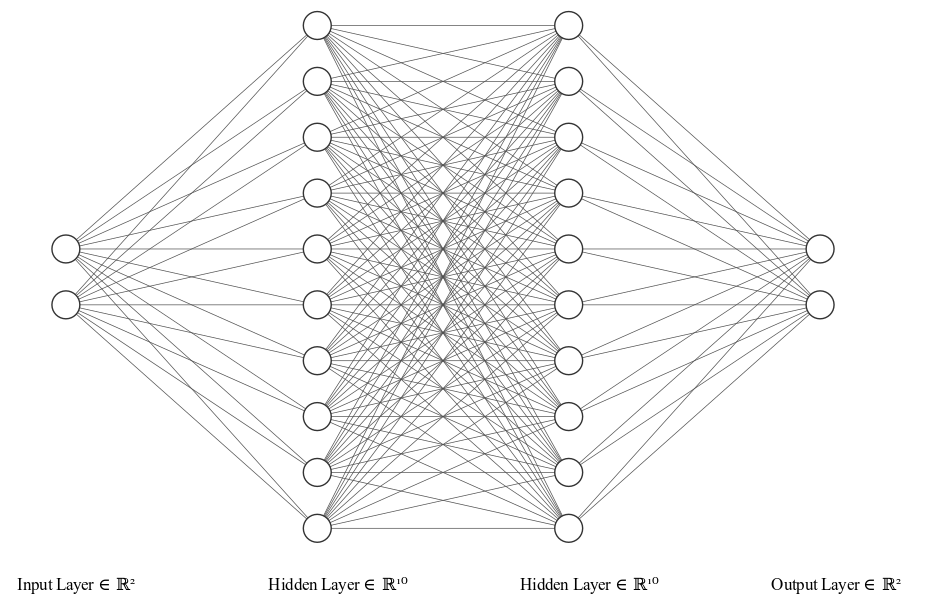}
        \caption{Depicts the Blobs architecture, it has two hidden layers with 10 ReLU neurons each.}
        \label{fig:topology-blobs}
  \end{minipage}
\end{figure}

We perform the experiments with two main types of NN architectures, denoted as \textit{XOR}, and \textit{Blobs}. Their topologies are depicted in Figures \ref{fig:topology-xor} and \ref{fig:topology-blobs} respectively. Two models were trained based on the \textit{XOR} topology (namely, models \textit{XOR-A} and \textit{XOR-B}), while one model was trained based on the \textit{Blobs} topology (denoted as \textit{Blobs} model). Each model was trained using a different dataset, while all are comprised of samples with the following shapes: $x \in \mathbb{R}^2$ and $y \in \{0,1 \}$. Therefore, our considered NNs were all trained for binary classification. 

About actual dataset generation, \textit{XOR-A}'s training data was generated by sampling points from a mixture of \textit{Gaussian} distributions, with centroids that correspond to the four points in $XOR_{mini}$ dataset (see \autoref{main:initial-exp}). Additionally, \textit{XOR-B}'s dataset was produced by re-sampling points from \textit{XOR-A}'s dataset. Specifically, blue points (i.e. with category value which is equal to 1) were given a lower probability for being selected. This down-sampling was required to assemble an unbalanced dataset, allowing us to explore a more realistic setting. Lastly, the \textit{Blobs}'s dataset was generated by sampling points from a mixture of \textit{Gaussian} distributions according to the following centroids, along with their labels: $\{(10, 10): 0, (-10, -10): 0, (-10, 10): 1, (10, -10): 1 , (30, -30): 0, (-30, -30): 1, (-30, 30): 0, (30, 30): 1\}$.

To evaluate our method, we measure the network accuracy before and after the repair, by using three sets of samples: (1) train set, (2) test set, and (3) uniformly sampled points (along with their corresponding classification according to the original network). Full details regarding the models and their corresponding datasets are summarized in Table \ref{tab:exp-setup}. Note that model \textit{XOR-A} is used to perform a sanity check experiment, and hence has no sampled set. For having a unified metric, we calculate a weighted average across all of the three sets with respect to their size. Reported experiments (\autoref{main:exp-compare}, \autoref{section:exp-repair}, \autoref{subsection:naive-baseline}) are evaluated based on this measure.

\begin{table}[htbp]
    \tiny
    \centering
    \setlength\tabcolsep{2pt}
    \begin{tabular}{|c|c|c|c|c|c|c|c|c|}
    \toprule
    \thead{Model \\ Name} & \thead{Hidden \\ Layers \\ Sizes} &  \thead{Parameters} &  \thead{Train \\ Set \\ Size} &  \thead{Train \\ Set \\ Accuracy} &  \thead{Test \\ Set \\ Size} &  \thead{Test \\ Set \\ Accuracy} &  \thead{Sampled \\ Set \\ Size} &  \thead{Sampled \\ Set \\ Accuracy} \\
    \midrule
    XOR-A & [4] &          22 &            2400 &             0.9991 &           1600 &            1.0 &               NA &                   NA \\
    XOR-B &           [4] &          22 &            1562 &             0.99743 &           1600 &            0.99125 &               500 &                   1.0 \\
    Blobs &      [10, 10] &         162 &            6000 &             1.00000 &           4000 &            1.00000 &               1000 &                   1.0 \\
    \bottomrule
    \end{tabular}
    \caption{Summarizes the examined NN models, together with their \textit{train}, \textit{test}, and \textit{sampled} sets statistics. \textit{XOR-A} and \textit{XOR-B} models were both originated from the same topology, though each was trained using a different dataset. The \textit{Blobs} model has its unique dataset as well, slightly larger than these of \textit{XOR-A} and \textit{XOR-B}. Also, note that the \textit{Blobs} model has a larger network topology and hence more parameters.}
    \label{tab:exp-setup}
\end{table}

\begin{figure}[!h]
    \captionsetup{font={small}}
	\begin{center}
    \makebox[\textwidth]{\includegraphics[width=0.75\paperwidth]{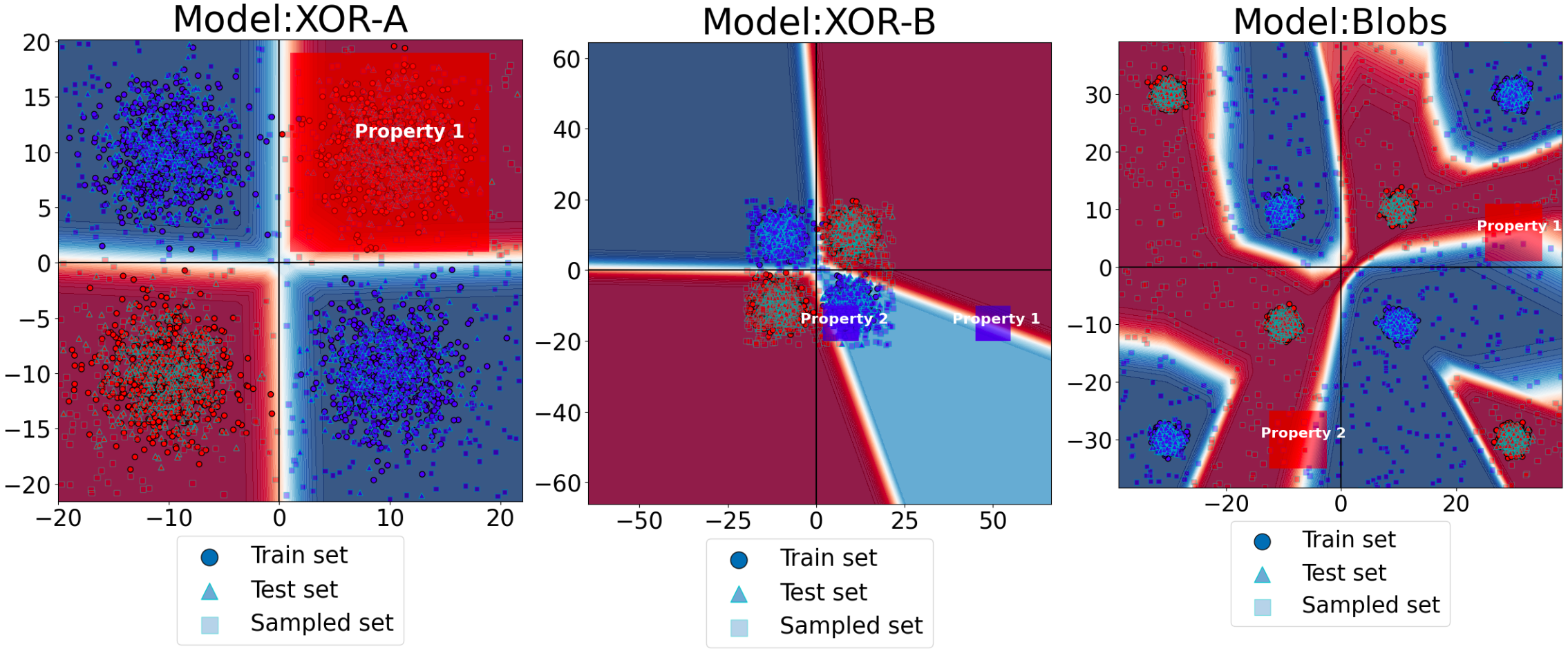}}
    \end{center}
    \caption{Depicts the three different models along with their decision boundaries, and the samples which assemble their data sets. Training samples correspond to circle shape points, test set samples have triangle shapes, and samples from the sampled set have a square shape.  Moreover, each of the model's safety properties is visualized as rectangles, and colored according to their desired category. Lastly, we note that model \textit{XOR-B}'s illustration shows zoomed-out boundaries since one of its properties is far from the training data. }
    \label{fig:models-properties}
\end{figure}

For each of our considered models, we assign the required safety properties. Figure \ref{fig:models-properties} illustrates the original model decision boundaries, including their desired specifications. First, model \textit{XOR-A} has its property ($Property^{XOR-A}_1$) already satisfied, as it is used for a sanity check. For model \textit{XOR-B} we select one property that is far from its training data ($Property^{XOR-B}_{1}$), and another one that is closer ($Property^{XOR-B}_2$), both properties are violated. Regarding the \textit{Blobs} model, we pick 2 properties ($Property^{Blobs}_1$ and $Property^{Blobs}_2$) relatively close to its training data, and in this case, both are violated. A detailed description of each property is depicted in Table \ref{tab:exp-specs}, note that this table uses the abbreviated notation of Property 1 and Property 2 for each considered model.

\begin{table}[htbp]
    \tiny
    \centering
    \setlength\tabcolsep{4pt}
    \begin{tabular}{|c|c|c|c|c|c|c|}
    \toprule
    \thead{Model \\ Name} & \thead{Property} &  \thead{Coordinate} &  \thead{Delta} & \thead{Class} & \thead{Norm} & \thead{Violated} \\
    \midrule
    XOR-A & Property 1 & [10, 10] & 9 & Red (0) & L1 & No \\
    XOR-B & Property 1 & [50, -15] & 5 & Blue (1) & L1 & Yes \\
    XOR-B & Property 2 & [7, -15] & 5 & Blue (1) & L1 & Yes \\
    Blobs & Property 1 & [30, 6] & 5 & Red (0) & L1 & Yes \\
    Blobs & Property 2 & [-7.5, -30] & 5 & Red (0) & L1 & Yes \\

    \bottomrule
    \end{tabular}
    \caption{Summarizes each of the considered models with their corresponding safety specifications. Model \textit{XOR-A} has one considered property which is satisfied. Models \textit{XOR-B} and \textit{Blobs} each have two required properties which are both violated, we attempt to repair these properties in our trials. All considered properties should enforce the adversarial robustness property, with respect to the \textit{L1} norm, their delta values are noted as well. Basically, these properties all form rectangle-shaped polygons as depicted in \autoref{fig:models-properties}}.
    \label{tab:exp-specs}
\end{table}

\subsubsection{Similarity Heuristics}
We consider a few heuristics to help with keeping the repaired model decisions as similar as possible to originals. In this section, we describe the mechanism behind each of these heuristics. All considered heuristics are implemented using soft constraints, thus, we intend to enable our solver to violate some of these constraints. Practically, when selecting a threshold value of 1 for soft constraints, the solver will attempt to satisfy \textbf{at least} 1 constraint. If it cannot satisfy even one constraint, then an UNSAT result is returned, otherwise, the solver returns SAT along with a solution (i.e., the new weight values). 

\paragraph{Samples:}\label{heuristics:samples}
Under the \textbf{Samples} heuristic, we enforce the model classification on certain coordinates.  More formally, each of these constraints has the form $x=x_0 \implies D(f_w(x)) = GT(x_0)$, meaning we require the model to classify point $x_0$ according to its correct label (i.e., ground truth). We select a subset of the NN training dataset to function as soft constraints for this heuristic. It is perhaps the cheapest heuristic to implement, and yet, we believe it is an effective heuristic as our initial experiments demonstrate. Through our trials, we use subsets of varying sizes for enforcing this heuristic.

\paragraph{Grid:}\label{heuristics:grid}
Using the \textbf{Grid} similarity heuristic, we explicitly encode the original decision boundary in a closed rectangle (or hyper-rectangle in the general case), by using a discrete grid to store its values. The grid limits are determined in correspondence to the training data boundaries. For each grid cell, we sample 3 points and take the NN majority vote as the required classification for this cell. There are several factors that might affect the results of this heuristic: (1) the number of cells and their sizes, (2) size and boundaries of the closed rectangle, (3) the number of samples (4) cells that intersect with boundary limits (i.e., areas in which the classification changes), etc. In an attempt to isolate the influence of these factors, we examined 3 separate configurations through our trials for this heuristic.

\begin{figure}
    \captionsetup{font={small}}
	\begin{center}
    \makebox[\textwidth]{\includegraphics[width=0.35\paperwidth]{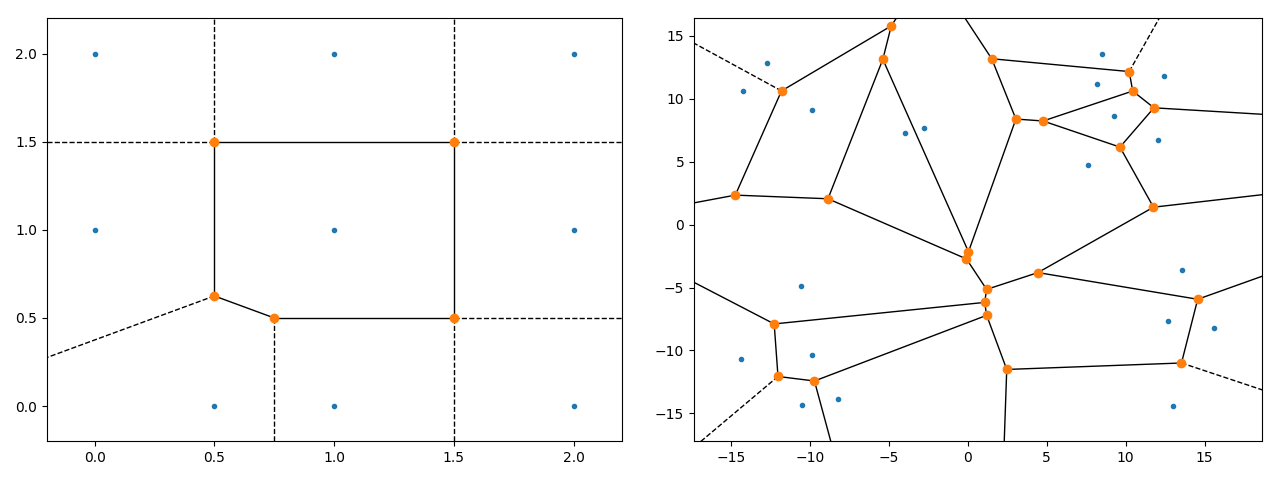}}
    \end{center}
    \caption{Depicts two distinct sets of samples (in blue) for generating Voronoi cells (with orange vertices). We can see how the sample's location and density affect the cells coverage.}
    \label{fig:heuristic-vor}
\end{figure}

\paragraph{Voronoi:}\label{heuristics:vor}
As another heuristic, inspired by works that involve Adversarial Defenses, we employ \textbf{Voronoi} tessellation (see \cite{fortune1995voronoi}) to produce an alternative to the grid constraints. Using this heuristic, we are given a finite set of points $\{p_1, ..., p_n\}$ in the Euclidean plane. Then, we build the cells according to the definition of Voronoi decomposition: for each point $p_k$ its corresponding Voronoi cell $R_k$ consists of every point in the Euclidean plane whose distance to $p_k$ is less than or equal to its distance to any other point $p_j$, where $j \neq k$.  Each cell category is determined by the label of its corresponding sample. Cells are built based on data points which were uniformly sampled from the original NN model (i.e., originated from the \textit{Sampled} dataset, see Table \ref{tab:exp-setup}). To affect the shapes and locations of resulting Voronoi cells, one could generate a unique subset of samples each time, Figure \ref{fig:heuristic-vor} demonstrates this phenomenon. In our trials, we inspect 4 different configurations for generating Voronoi cells, that is by re-sampling points out of the \textit{Sampled} dataset, to yield diverse shapes of Voronoi cells for the same problem. 

To formulate the actual constraints for utilizing this heuristic, we employ the \textit{SciPy} Python package \cite{2020SciPy-NMeth} which allows constructing Voronoi regions, as illustrated in \autoref{fig:heuristic-vor}. Then, each of these regions are transformed into a conjunction of linear equations.

\subsubsection*{Technical Details}
Throughout the trials of further described experiments\footnote{Code base and results are available at https://github.com/dorcoh/NNSynthesizer}, we use Z3, as in our initial experiments. To train the networks, we employ Adam optimizer \cite{kingma2017adam} with an NLL loss function, using PyTorch \cite{paszke2017automatic} framework. To compute Voronoi cells, we assist with SciPy Python package \cite{2020SciPy-NMeth}. Experiments were performed on a computer with an Intel(R) Core(TM) i5-8250U CPU @ 1.60GHz using 8Gb of RAM, running Linux Ubuntu version 16.04.1. Each trial was given a timeout of 600 seconds.

\subsection{Effecting the result by explicit restriction of data points}\label{main:initial-exp-degree}
\begin{figure}
	\begin{center}
    \makebox[\textwidth]{\includegraphics[width=0.75\paperwidth]{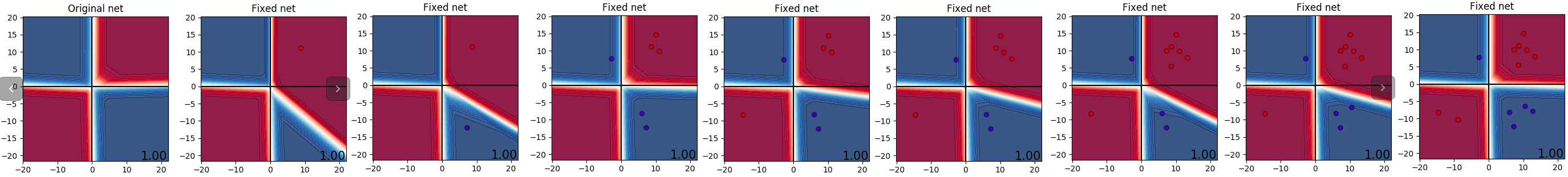}}
    \end{center}
	\caption{Depicted on the most left figure is the original decision boundary for \textit{XOR-A}, and repaired boundaries for the rest of the figures. In this experiment, we attempt to repair an already satisfied property. Each repair trial utilizes an increasing amount of similarity-preserving constraints, starting from the second figure from the left with a single additional constraint, and up to the rightmost figure with 13 additional constraints. We can observe how utilizing a varying number of constraints lead to relatively diverse results.}
	\label{fig:initial-exp-degree}
\end{figure}

First, to affect the result of our SMT solver, we attempt to enforce a various number of additional similarity-preserving constraints. Our goal is to demonstrate that these types of constraints assist with retaining the original boundaries. For this purpose, our candidate network is XOR-A, while the employed similarity heuristic is \textit{Samples} (\autoref{heuristics:samples}). We free one weight during the search, additionally, in our formulation we seek to repair a property that is already satisfied. In Figure \ref{fig:initial-exp-degree} we can see how increasing the number of constraints results in more similar boundaries, however, the improvement rate is not always consistent. It is possible that the number of total constraints is not sufficient, therefore we should perhaps use a larger set in further trials. Generally, this experiment validates that our method can yield diverse results when utilizing a varying number of constraints.

\subsection{Experiments - Compare Similarity Heuristics}\label{main:exp-compare}

We start with a comparative study of our considered similarity-preserving heuristics. In this experiment, we repair model \textit{XOR-B} w.r.t. its safety properties, by utilizing each of the heuristics. For each property, we perform the search using slightly different weights selection tactics as follows: (1) enumerate over all individual weights, (2) arbitrarily pick combinations of weights, neurons, or layers (i.e., select all neurons in a layer). The goal of this experiment set is to assess the effectiveness and efficiency of our examined similarity heuristics. For this purpose, in each trial, we record various accuracy measures (measured on train/test/sample sets) and report weighted average across these sets (as described earlier in \autoref{main:exp-setup}). We also measure the solver time to assess how efficient are these heuristics in practice.

Table \ref{tab:compare-heuristics} depicts aggregated metrics from this experiment. Specifically, for each examined heuristic, this table summarizes statistics (i.e., max, min, and average accuracy after the repair process) for the unified accuracy measure (see \autoref{subsection::setup}) across all trials, number of trials, number of successful repairs (i.e., the solver returns SAT), unsuccessful repairs (UNSAT), and trials which were resulted in a timeout. Lastly, we also recorded the total solver time in seconds for each group of trials. 

For the first property, the best result in terms of max accuracy ($98.60\%$) is achieved by selecting a combination of arbitrary weights, and by using the \textit{Samples} heuristic. For the second property, combinations of individual weights were sufficient to achieve an accuracy rate of $98.98\%$ using both \textit{Samples} or \textit{Grid} heuristics. To note, in all of the trials where \textit{Grid} or \textit{Voronoi} heuristics were employed, the SMT solver found no solutions (i.e., resulted with UNSAT or Timeout) beyond threshold values which are larger than 1. The results generally demonstrate that the \textit{Samples} heuristic is more precise on average, in terms of accuracy. 

Moreover, we observe that trials which employed \textit{Grid} or \textit{Voronoi} heuristics can take a longer time to compute when compared to \textit{Samples} heuristic trials. Likewise, the number of trials which resulted in timeout is relatively smaller when using the \textit{Samples} heuristic. Additionally, as could be expected, it is evident that selecting single weights results in a faster search when compared to searching over multiple free weights. Perhaps this observation can be utilized to perform a kind of greedy search. For instance, at the first stage, we free single weights and get the results relatively fast. Then, combinations of two or more weights can be assembled out of the top performing trials in which we freed single weights. 

\begin{table}[htbp]
    \tiny
    \centering
    \setlength\tabcolsep{2pt}
    \begin{tabular}{|c|c|c|c|c|c|c|c|c|c|c|c|}

    \toprule
    \thead{Heuristic} & \thead{Soft \\ Constraints} &    \thead{Thresholds} &  \thead{Accuracy \\ Before \\ Repair} &   \thead{Max. \\ Accuracy } &   \thead{Min. \\ Accuracy} &   \thead{Average \\ Accuracy} & \thead{Trials} & \thead{SAT} &  \thead{UNSAT} &  \thead{Timeout} &  \thead{Total \\ Solver \\ Time \\(sec.)} \\
    \midrule
    
    \multicolumn{12}{|c|}{\textbf{Individual weights}} \\
    \hline
    
    \multicolumn{12}{|c|}{\nth{1} Property} \\
    \midrule
     GRID &           [100] &        [1, 2] &   99.5085\% & 85.9640\% & 37.5205\% & 60.1614\% & 44 &    9 &     26 &        9 &             6397 \\
  SAMPLES &           [450] & [1, 250, 400] &   99.5085\% & 86.9470\% & 37.5205\% & 71.8988\% & 66 &   15 &     51 &        0 &               32 \\
  VORONOI &           [153] &        [1, 2] &   99.5085\% & 85.9640\% & 37.5205\% & 55.9803\% & 44 &    9 &     13 &       22 &            13994 \\
    
    \midrule
    \multicolumn{12}{|c|}{\nth{2} Property} \\
    \midrule
    
     GRID &         [100] &        [1, 2] &   99.5085\% & $\boldsymbol{98.9896\%}$ & 37.5205\% & 65.5980\% & 44 &    10 &     24 &       10 &             6234 \\
  SAMPLES &         [450] & [1, 250, 400] &   99.5085\% & $\boldsymbol{98.9896\%}$ & 37.5205\% & 83.6997\% & 66 &   24 &     42 &        0 &               32 \\
  VORONOI &         [153] &        [1, 2] &   99.5085\% & 96.3681\% & 37.5205\% & 60.6827\% &  44 &    10 &     12 &       22 &            13843 \\
    
    \midrule
    \multicolumn{12}{|c|}{\textbf{Arbitrary combinations of weights, neurons or layers}} \\
    \midrule
    
    \multicolumn{12}{|c|}{\nth{1} Property} \\
    \midrule
     GRID &      [100, 400] &    [1, 2, 10] &   99.5085\% & 85.9640\% & 37.5205\% & 60.9448\% &  38 &  15 &      0 &       23 &            13812 \\
  SAMPLES &          [450] & [1, 250, 400] &   99.5085\% & $\boldsymbol{98.6073\%}$ & 37.5205\% & 84.4578\% &  36 &     32 &      4 &        0 &              171 \\
  VORONOI &[268, 250, 153] &    [1, 2, 10] &   99.5085\% & 86.2370\% & 37.5205\% & 60.9030\% & 39 &    15 &      0 &       24 &            14424 \\
    \midrule
    \multicolumn{12}{|c|}{\nth{2} Property} \\
    \midrule
     GRID &     [100, 441] &             [1, 2] &   99.5085\% & 59.3392\% & 37.5205\% & 44.1168\% &  55 &    9 &     27 &       19 &            11522 \\
  SAMPLES &          [450] & [1, 250, 400, 450] &   99.5085\% & 98.0339\% & 37.5205\% & 83.7520\% & 56 &    13 &     36 &        7 &             4339 \\
  VORONOI &      [153, 82] &             [1, 2] &   99.5085\% & 80.2567\% & 37.3020\% & 47.9465\% &  56 &   10 &     18 &       28 &            16869 \\
    \bottomrule
    \end{tabular}
    \caption{Aggregated results of our first set of experiments in which we compare similarity-preserving heuristics, by repairing \textit{XOR-B}'s safety properties. We employ here two weight-selection tactics: (1) selecting all individual weights (results depicted at the upper block of this table), (2) arbitrarily picking combinations of weights (lower block). In all trials, \textit{Samples} heuristic proves to be the most effective and efficient, as measured with accuracy and solver time respectively. Likely, \textit{Grid} and \textit{Voronoi} heuristics get inferior but comparable results in terms of accuracy.}
    \label{tab:compare-heuristics}
\end{table}

\begin{figure}
    \captionsetup{font={small}}
	\begin{center}
    \makebox[\textwidth]{\includegraphics[width=0.8\paperwidth]{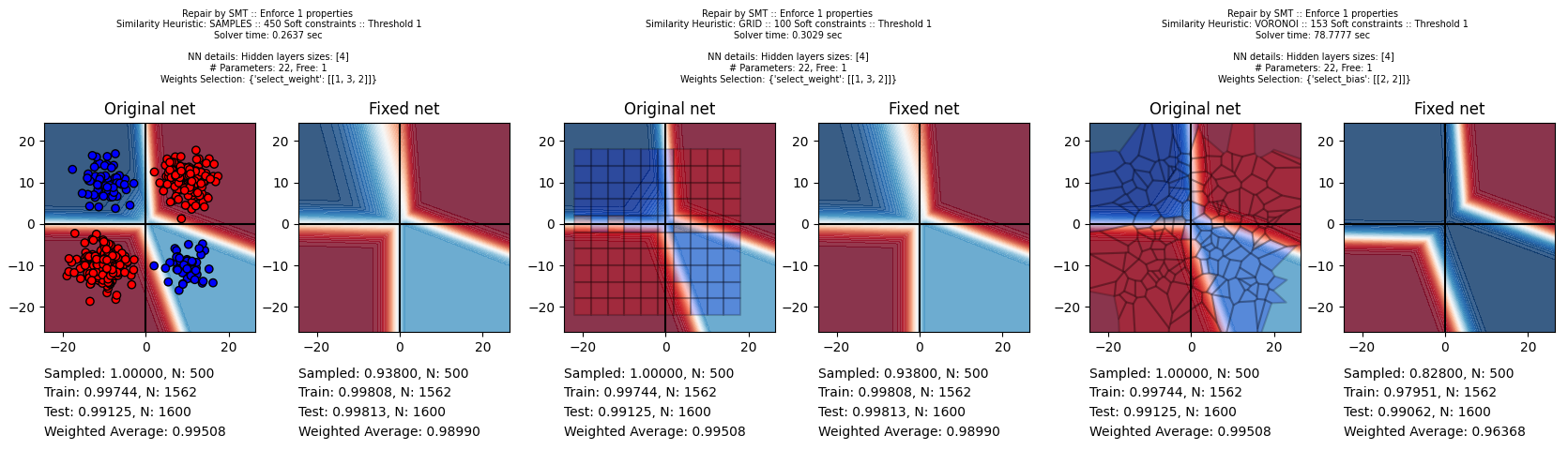}}
    \end{center}
    \caption{Illustrates the decision boundaries of the best three repair trials, for each of the considered heuristics, when attempting to repair $Property^{XOR-B}_{1}$. Each pair of figures relates to one trial performed with a different heuristic (located from left to right: \textit{Samples}, \textit{Grid}, and \textit{Voronoi} heuristics). The left figure in each pair depicts the original network, along with a sketch of the similarity heuristic, while the right figure depicts the repaired boundaries. Below each pair, depicted are the accuracy measures before and after the repair on the left and right sides respectively. Lastly, additional details regarding the trial are depicted above each pair.}
    \label{fig:exp1-merged}
\end{figure}

To visually demonstrate our method's effectiveness, we pick the best result for each heuristic and show the network boundaries before and after the repair. Figure \ref{fig:exp1-merged} depicts the decision boundaries before and after repairing the network \nth{1} property (i.e., $Property^{XOR-B}_{1}$). We can observe how all of these trials were successfully completed with a threshold value of 1. In essence, a threshold of 1 allows the solver to pick its "favored" number of satisfied constraints. Additionally, in all of these trials, only a single weight is free. Interestingly, in this trial, both \textit{Samples} and \textit{Grid} heuristics yield the same result (i.e., identical weight values). 

To summarize, as follows from this experiment, our examined similarity-preserving heuristics seem to be comparable to each other in terms of effectiveness, with a slight advantage to \textit{Samples} heuristic. However, when considering their efficiency (i.e., as resembled from the sum of solver time), then the \textit{Samples} heuristic significantly outperforms \textit{Grid} and \textit{Voronoi} heuristics.

\subsection{Experiments - Repair with \textit{Samples} Similarity Heuristic}\label{section:exp-repair}

In this section, we proceed with the task of repairing networks \textit{XOR-B} and \textit{Blobs} by considering only the \textit{Samples} similarity heuristic, which was empirically proved to be the most efficient, as demonstrated in the previous section. We then compare the results of repairing these networks to a naive baseline. Briefly, this baseline implements a repair loop that alternates between verifying the network safety and re-training it using additional data points. We give more details about this comparison in \autoref{subsection:naive-baseline}. In the following, we thoroughly explain the attempts to repair these networks with our method.

Similar to the previous section (\autoref{main:exp-compare}), aggregated results are presented in \autoref{tab:results-xorb} and \autoref{tab:results-blobs}. In this experiment set, however, we group the results according to the considered threshold value. The intention is to understand whether the threshold value is correlated with other metrics. Moreover, we employ a skip mechanism as follows - for each trial, we try to increase the threshold value (e.g., 1, 325, 750, etc.) as long as the solver returns SAT. Once the solver returns UNSAT or it has timed out, we skip further configurations. That is the reason why some results were not available (marked as NA in \autoref{tab:results-xorb} and \autoref{tab:results-blobs}).

Regarding weight selection, the \textit{XOR-B} network has $22$ parameters and $231$ combinations of weight pairs. While \textit{Blobs} network has $162$ parameters and $13041$ potential pairs. Therefore, in several trials, we use a simple weight selection tactic (i.e., select all combinations of single weights, or all pairs of weights), wherein others we employ a greedy weight selection tactic (i.e., assemble weight pair combinations out of top-performing attempts with single weights).

\begin{table}[htbp]
    \tiny
    \centering
    \setlength\tabcolsep{2pt}
    \begin{tabular}{|c|c|c|c|c|c|c|c|c|c|c|}

    \toprule
       \thead{Threshold / Soft \\ Constraints} & \thead{Accuracy \\ Before \\ Repair} &    \thead{Max. \\ Accuracy} &    \thead{Min. \\ Accuracy} &    \thead{Average \\ Accuracy} & \thead{Trials} &   \thead{SAT} &  \thead{UNSAT} &  \thead{Timeout} &  \thead{Skipped} &  \thead{Total \\ Solver \\ Time \\ (sec.)} \\
    \midrule
    
    \multicolumn{11}{|c|}{\textbf{1st Property}} \\ 
    \midrule
    
    \multicolumn{11}{|c|}{Individual weights} \\
    \midrule
    1/1562 & 99.50847\% & 85.96395\% & 37.52048\% & 66.35718\% &  22 &     9 &     13 &        0 &        0 &              110 \\
    325/1562 & 99.50847\% & 84.21628\% & 37.52048\% & 64.60343\% & 22 &      9 &     13 &        0 &        0 &               70 \\
    750/1562 & 99.50847\% & 85.96395\% & 49.56308\% & 70.17340\% & 22 &     8 &     14 &        0 &        0 &               69 \\
    1000/1562 & 99.50847\% & 86.94702\% & 84.21628\% & 85.58165\% & 22 &     4 &     18 &        0 &        0 &               57 \\
    1500/1562 & 99.50847\%  &        NA &        NA &        NA & 22 &     0 &     22 &        0 &        0 &               58 \\
    1561/1562 & 99.50847\%  &        NA &        NA &        NA & 22 &     0 &     22 &        0 &        0 &               34 \\
    
    \midrule
    \multicolumn{11}{|c|}{Combinations of two weights} \\
    \midrule
    
    1/1562 &  99.50847\% & 99.20808\% & 33.72474\% & 65.91046\% & 231 &   156 &     61 &       14 &        0 &             9127 \\
    325/1562 &  99.50847\% & 99.12616\% & 37.73894\% & 68.05714\% & 231 &   103 &      0 &       53 &       75 &            42027 \\
    750/1562 &  99.50847\% & $\boldsymbol{99.26270\%}$ & 48.90770\% & 71.35950\% & 231 &    92 &      0 &       11 &      128 &            18004 \\
    1000/1562 &  99.50847\% & 98.60732\% & 61.22338\% & 80.55633\% & 230 &    74 &     11 &        6 &      139 &            13321 \\
    1500/1562 &  99.50847\% & $\boldsymbol{99.26270\%}$ & 93.11851\% & 96.96887\% & 230 &  18 &     51 &        5 &      156 &             9529 \\
    
    \midrule
    \multicolumn{11}{|c|}{\textbf{2nd Property}} \\
    \midrule
    \multicolumn{11}{|c|}{Individual weights} \\
    \midrule
    1/1562 & 99.50847\% & 98.98962\% & 38.09394\% & 74.23539\% &  22 &      10 &     12 &        0 &        0 &              141 \\
    325/1562 & 99.50847\% & 97.95194\% & 37.52048\% & 74.13435\% &  22 &      10 &     12 &        0 &        0 &               75 \\
    750/1562 & 99.50847\% & 97.95194\% & 49.80885\% & 77.06475\% &   22 &     9 &     13 &        0 &        0 &               63 \\
    1000/1562 & 99.50847\% & 97.95194\% & 64.09066\% & 89.54806\% &   22 &    8 &     14 &        0 &        0 &               68 \\
    1500/1562 & 99.50847\% & $\boldsymbol{99.31731\%}$ & 96.36810\% & 97.50137\% &   22 &    4 &     18 &        0 &        0 &               64 \\
    1561/1562 & 99.50847\% &        NA &        NA &        NA &  22 &     0 &     22 &        0 &        0 &               27 \\
    
    \midrule
    \multicolumn{11}{|c|}{\textbf{1st And 2nd Property}} \\
    \midrule
    \multicolumn{11}{|c|}{Individual weights} \\
    \midrule
    
    1/1562 & 99.50847\% & 84.21628\% & 37.52048\% & 61.06499\% &  22 &     5 &     17 &        0 &        0 &              119 \\
    325/1562 & 99.50847\% & 84.21628\% & 37.52048\% & 61.04315\% &  22 &     5 &      0 &        0 &       17 &               31 \\
    750/1562 & 99.50847\% & 84.21628\% & 49.45385\% & 66.92381\% &  22 &     4 &      1 &        0 &       17 &               27 \\
    1000/1562 & 99.50847\% & 84.21628\% & 84.21628\% & 84.21628\% & 22 &      2 &      2 &        0 &       18 &               16 \\
    1500/1562 & 99.50847\% &        NA &        NA &        NA & 22 &     0 &      2 &        0 &       20 &                7 \\
    1561/1562 & 99.50847\% &        NA &        NA &        NA & 22 &     0 &      0 &        0 &       22 &                0 \\
    
    \midrule
    \multicolumn{11}{|c|}{Combinations of two weights} \\
    \midrule

    1/1562 &  99.50847\% & 97.21464\% & 20.04369\% & 57.71723\% &  231 &   119 &     95 &       14 &        0 &             8933 \\
    325/1562 &  99.50847\% & 90.44238\% & 36.15511\% & 63.79459\% & 231 &    76 &      0 &       43 &      112 &            34731 \\
    750/1562 &  99.50847\% & $\boldsymbol{97.35117\%}$ & 48.90770\% & 68.24854\% & 231 &    65 &      0 &       11 &      155 &            11555 \\
    1000/1562 &  99.50847\% & 90.44238\% & 63.62643\% & 80.96122\% & 231 &    40 &     16 &        9 &      166 &            11202 \\
    1500/1562 &  99.50847\% & 97.05079\% & 96.28618\% & 96.66849\% &  231 &    2 &     37 &        1 &      191 &             3676 \\
    
    \bottomrule
    \end{tabular}
    \caption{Aggregated search results for the repair of \textit{XOR-B} network, with \textit{Samples} similarity heuristic, results are grouped by soft constraints threshold value. This table demonstrates the capability of our framework to produce safe networks by freeing up to two weights, with only a mild loss of accuracy. }
    \label{tab:results-xorb}
\end{table}


\autoref{tab:results-xorb} depicts the repair results when attempting to repair the \textit{XOR-B} network and its properties. For its first property, the best performing trial with an accuracy of $99.26\%$ uses a combination of 2 weights, with a threshold of 750 out of 1562 soft constraints. Surprisingly, the same result is achieved also with a threshold of 1500. When using combinations of 1 weight, the best result our method achieves is $86.94\%$. Thus, by freeing another weight we improve the error rate by $1670\%$. This observation implies that our framework might achieve even better results, by considering combinations sets of larger sizes (i.e., more than 2 weights). About \textit{XOR-B}'s second property, we experimented with combinations of single weights only, and reached the best result with an accuracy $99.31\%$, by setting a threshold of at least 1500 constraints. Finally, when attempting to repair the two properties, the best result with an accuracy of $97.35\%$ is achieved by a combination of two weights and a threshold of at least 750 constraints.  Furthermore, it is evident that increasing the threshold value yields on average more similar representations (in terms of average accuracy across all group trials). It is also clear that freeing a pair of weights instead of single weights heavily affects the solver time, perhaps a longer timeout should be considered.

\begin{table}[htbp]
    \tiny
    \centering
    \setlength\tabcolsep{2pt}
    \begin{tabular}{|c|c|c|c|c|c|c|c|c|c|c|}

    \toprule
      \thead{Threshold / \\ Soft \\ Constraints} & \thead{Accuracy \\ Before \\ Repair} &    \thead{Max. \\ Accuracy} &    \thead{Min. \\ Accuracy} &    \thead{Average \\ Accuracy} &  \thead{Trials} &  \thead{SAT} &  \thead{UNSAT} &  \thead{Timeout} &  \thead{Skipped} &  \thead{Total \\ Solver \\ Time \\ (sec.)} \\
    \midrule
    \multicolumn{11}{|c|}{\textbf{1st Property} (individual weights)} \\ 
    \midrule
    
    1/6000 & 100.00000\% & 99.43636\% & 64.10000\% & 89.52500\% & 107 &   48 &     44 &       15 &        0 &            18968 \\
    500/6000 & 100.00000\% & 99.41818\% & 86.50909\% & 95.03438\% & 107 &   32 &      0 &       16 &       59 &            11723 \\
    1000/6000 & 100.00000\% & $\boldsymbol{99.50000\%}$ & 86.50909\% & 94.11420\% & 107 &    32 &      0 &        0 &       75 & 2046 \\
    2000/6000 & 100.00000\% & 99.43636\% & 86.47273\% & 94.44403\% & 107 &    32 &      0 &        0 &       75 &             2330 \\
    5500/6000 & 100.00000\% & 99.34545\% & 90.49091\% & 95.55016\% & 107 &   29 &      3 &        0 &       75 &             3518 \\
    
    \midrule
    \multicolumn{11}{|c|}{\textbf{2nd Property} (individual weights)} \\
    \midrule
    
    1/6000 & 100.00000\% & 99.72727\% & 51.12727\% & 86.66129\% &  143 &    31 &     95 &       17 &        0 &            23884 \\
    500/6000 & 100.00000\% & 99.72727\% & 86.15455\% & 92.72323\% &  143 &   18 &      0 &       13 &      112 &             9573 \\
    1000/6000 & 100.00000\% & 99.72727\% & 86.30909\% & 93.59798\% & 143 &    18 &      0 &        0 &      125 &             1406 \\
    2000/6000 & 100.00000\% & $\boldsymbol{99.75455\%}$ & 64.67273\% & 90.52121\% & 143 &    18 &      0 &        0 &      125 &  1789 \\
    5500/6000 & 100.00000\% & 99.63636\% & 90.47273\% & 94.50160\% & 143 &    17 &      1 &        0 &      125 &             2580 \\
    
    \midrule
    \multicolumn{11}{|c|}{\textbf{1st And 2nd Property}} \\ 
    \midrule
    \multicolumn{11}{|c|}{Combinations of size 2, assembled from top-5 repair setups of individual weights} \\
    \midrule
    1/6000 & 100.00000\% & 99.49091\% & 62.45455\% & 89.15565\% &  45 &    33 &     12 &        0 &        0 &             5137 \\
    500/6000 & 100.00000\% & 99.10909\% & 74.16364\% & 93.26263\% &  45 &     9 &      0 &       24 &       12 &            15823 \\
    1000/6000 & 100.00000\% & $\boldsymbol{99.56364\%}$ & 86.30909\% & 93.08636\% &  45 &     4 &      0 &        5 &       36 &  3477 \\
    2000/6000 & 100.00000\% & 86.89091\% & 86.30909\% & 86.55455\% &  45 &    3 &      0 &        1 &       41 &              747 \\
    5500/6000 & 100.00000\% & 98.81818\% & 97.35455\% & 98.23636\% &  45 &    3 &      0 &        0 &       42 &              638 \\
    
    \bottomrule
    \end{tabular}
    \caption{Aggregated search results for the repair of \textit{Blobs} network, with \textit{Samples} similarity heuristic, results are grouped by soft constraints threshold value. This table demonstrates that it might be possible to exploit previous search results to produce a greedy heuristic for traversing the considered search tree. Specifically, the information from repairing single properties might be useful to further attempts of repairing multiple properties simultaneously.}
    \label{tab:results-blobs}
\end{table}

\autoref{tab:results-blobs} Summarizes the repair results for the \textit{Blobs} network. In this experiment, we employ a greedy weight selection tactic when attempting to repair both properties simultaneously. Specifically, we first attempt to repair each property separately by freeing single weights. Then, we pick the top 5 repair results and select their free weights as candidates for assembling combinations of size 2. This greedy search technique proves to achieve reasonable performance with a decreased compute overhead. It can also be extended to search for combinations of larger sizes.

Regarding \textit{Blobs}'s first property, the most similar network in terms of accuracy is achieved by employing at least 1000 out of 6000 soft constraints, with an accuracy rate of $99.50\%$. The second property is best repaired with a threshold of 2000 constraints, its network achieves an accuracy of $99.75\%$. While for both properties, the best threshold is equal to 1000 constraints with an accuracy of $99.56\%$. It is noticeable that the solver time in this experiment is an order of magnitude larger than the previous one (\autoref{tab:results-xorb}), apparently due to the enlarged network size (i.e., 162 vs. 22 parameters). Further, we can observe how the threshold value is less correlated with the average accuracy when compared to the previous experiment (\autoref{tab:results-xorb}). This observation suggests that perhaps there is a threshold value that is sufficient for reaching the best accuracy. Therefore, we might search for this value prior to the repair procedure, in another attempt to save compute time.


\subsubsection{Comparison to baseline}\label{subsection:naive-baseline}

As noted, we compare our proposed method with a naive one - a gradient-based search repair loop. The goal here is to demonstrate that our method is comparable to a slightly simple but effective baseline. Specifically, this baseline implements a repair loop that optimizes the network weights via traditional gradient-based methods. More concretely, this loop queries an SMT solver for verifying the NN safety. Then, as long as the NN is unsafe, it samples new data points out of the given specification and uses them as additional training points to retrain the network. We exit this loop when the network is safe, or when it reaches a certain number of iterations. We set a limit of 20 such iterations for convergence. Our sampling method attempts to take into account the balance between points originating from the training data, versus points that were sampled from the desired specification. Adding more samples from the latter group shall lead to faster convergence (i.e., a SAT result), although it may also produce less precise networks in terms of accuracy. Thus, at each resampling iteration, we extend the training set by 200 new points which originated from the desired specification, and 200 new points sampled from the original training set.

\begin{table}[htbp]
    \tiny
    \centering
    \setlength\tabcolsep{2pt}
    \begin{tabular}{|c|c|c|c|c|c|c|}
    \hline
    Model        & \multicolumn{3}{c|}{XOR-B}                                                                  & \multicolumn{3}{c|}{Blobs}                                                           \\ \hline
    Property     & \multicolumn{1}{c|}{1st}         & \multicolumn{1}{c|}{2nd}  & \multicolumn{1}{c|}{Both}  & \multicolumn{1}{c|}{1st}   & \multicolumn{1}{c|}{2nd}   & \multicolumn{1}{c|}{Both} \\ \hline
    Our method   & \textbf{99.2627\%} & \textbf{99.3173\%}  & \textbf{97.3511\%} & \textbf{99.5000\%} & \textbf{99.7545\%} & \textbf{99.5636\%} \\ \hline
    Naive method & 97.2678\% & 96.4469\% & 94.5724\% & 97.3533\% & 97.8333\% & 96.0733\% \\ \hline
    \end{tabular}
    \caption{Comparing our method to a naive baseline for repairing the network, depicted are the reported accuracy measures.  We observe that our method outperforms the proposed naive method in terms of accuracy, across all tested configurations.}
    \label{tab:compare}
\end{table}

The results in \autoref{tab:compare} confirm that our method is comparable to a naive baseline. In all trials, our method outperforms the baseline in terms of effectiveness (as measured by accuracy). However, we also note that using our method is much more expensive in terms of computation. In this experiment, the naive repair loop lasts a few tens of seconds at most. On the other hand, our method could take tens of thousands of seconds, especially when freeing more than 1 weight as depicted in \autoref{tab:results-xorb} and \autoref{tab:results-blobs}. Nevertheless, we expect both methods to quickly become computationally harder when considering larger networks. Additionally, we note that the repair process with our method is composed of tens of trials, against the baseline, which is the result of one trial for each configuration. Therefore, it is very likely that the baseline can be further tuned. For example, one factor which could affect the results is the ratio of the original training data points versus the points which were sampled from the specification. Tuning this ratio might assist to produce more precise networks, at the cost of extra computation.
One clear distinction between our method and the baseline is the number of free weights. Gradient-based methods search across all network parameters by utilizing a loss function (i.e., the objective function) to provide 'direction' through the search. While in our method we free only a small subset of the network weights and use heuristics for precision. However, as this experiment shows, this fact might help our method to provide more precise results in terms of similarity to the original network. 

\subsection{Automated Repair of Neural Networks}
\label{section:proposed-alg}

Our experiments demonstrate that using our method, it might be reasonable to perform a greedy search. That is, we do not necessarily need to enumerate all possible weight combinations. Specifically, we found that for some weight combinations of size 1, the solver returns UNSAT, while for others SAT is returned. Therefore, these results can be further utilized to build a search tree. Nodes in this tree resemble free weight combinations, while the level of the tree indicates the size of these combinations (i.e., the number of free weights). Then, according to the solver results, large parts of this search tree could be eliminated. For instance, we can decide to eliminate any paths that include nodes which resulted in UNSAT. Intuitively, this branching heuristic should assist with getting fewer UNSAT results through the search process. In the following, we describe a sketch for a greedy search method that takes advantage of this observation.

\begin{algorithm}
\caption{Greedy repair}
\begin{algorithmic}[1]
\REQUIRE{$nn \leftarrow NeuralNetwork$, \\ $spec \leftarrow Specification$, \\ $timeout \leftarrow TrialTimeout$, \\ $global\_timeout \leftarrow GlobalTimeout$ \\}
\hrulefill
\STATE{$best\_result \leftarrow RepairResult()$}
\FOR{$size = 1$; $size < nn.number\_of\_weights $; $size \leftarrow size+1$}
    \STATE{$combinations \leftarrow$ BuildEligibleCombinations($results,size$)}
    \FOR{$comb$ in $combinations$}
        \STATE{$result \leftarrow$ Solve($nn, spec, comb, timeout$)}
        \IF{$result.return\_value == SAT$ $\AND$ $result.accuracy > best\_result.accuracy$}
            \STATE{$best\_result \leftarrow RepairResult(result)$}
        \ENDIF
    \IF{ExitConditionIsTrue($global\_timeout$)}
        \RETURN{$best\_result$}
    \ENDIF
    \ENDFOR
\ENDFOR
\RETURN{$best\_result$}
\end{algorithmic}
\end{algorithm}
\label{alg:first}

Our greedy optimization method is depicted in \autoref{alg:first}. First, it takes as input the neural network representation (denoted as $NeuralNetwork$), its desired specification (denoted as $Specification$), and timeout values in seconds (denoted as $TrialTimeout$ for the solver timeout and $GlobalTimeout$ for the total elapsed time). On each iteration, it builds the next valid combinations (by calling $BuildEligibleCombinations$) according to the current status of the search tree. Then, the SMT solver is invoked (denoted as $Solve$) to check each of these configurations. After each trial, a check is performed and the loop halts if the elapsed time exceeds our desired timeout (this function is noted as $ExitConditionIsTrue$). At this point, we return the best result.

On each level of the search (denoted by $size$ in \autoref{alg:first}), $BuildEligibleCombinations$ produces ${n \choose k}$ combinations in the worst case, where $n$ denotes the total number of weights, and $k$ refers to the current level (i.e., $size$). By expanding the term $O({n \choose k})$ and dividing it into cases: (1) $k <= n - k$, (2) $n - k < k$, it ends up being equivalent to $O(n^{min\{k, n-k\}})$, for a constant $k$ value. Hence, the complexity to enumerate through all combinations is $\sum_{k=1}^{n} O(n^{min\{k, n-k\}})$, which is equal to $O(n^{\frac{n}{2}})$, since the largest element of this sum is achieved at $k=\frac{n}{2}$. As this analysis suggests, we might want to use heuristics for deciding how to branch on each stage of the search tree. Thus, we provide the following heuristic, as follows from our experiments. 

\paragraph{Branching heuristic.} On each level scanning the search tree, UNSAT weight combinations are marked, and we no longer consider new trials which involve these combinations. Therefore, by using this heuristic, \autoref{alg:first} solution (best result) is assembled from the weights which their corresponding trials resulted in SAT.

For example, assuming that after scanning the 1st level, the following combinations were found to yield SAT and UNSAT results, respectively: $\{w_{111}, w_{112}, w_{113}\}$, $\{w_{114}\}$. Then, on the second level, we consider only these configurations: $\{ \{w_{111}, w_{112}\}, \{w_{111}, w_{113}\}, \{w_{112}, w_{113}\}\}$. Naturally, by following this mechanism, some valid configurations can be omitted. For instance, the combination $\{w_{113}, w_{114}\}$ will not be checked under this heuristic, even though it might return SAT. Hence, the intention behind this heuristic is to try and reduce the total exploration time, that is, by skipping combinations which might lead to UNSAT results.

Note that our algorithm can be also further improved by taking advantage of parallelism. Clearly, given sufficient resources, the solver can be executed (i.e., by calling $Solve$) multiple times simultaneously. Through our trials, we perform each such computation separately, then we aggregate the results to get the best one.

As described above, we might want to improve the algorithm performance and efficiency, then we propose a few more strategies which can be further explored:

\begin{itemize}
  \item Change the exit condition as defined by $ExitConditionIsTrue$ (for example, to take into account the desired accuracy)
  \item Search certain locations of the network topology. For instance, we can replace only weights in the last layer, considerably reducing the number of combinations. (this technique for optimizing deeper layers only is common among ML practitioners and is also known as 'Transfer Learning' (\cite{tan2018survey}). Its purpose is to use the knowledge gained in solving one problem in another related but different problem.)
  \item Search for the soft constraints threshold value. Observe that we do not search for the threshold in our current algorithm, instead, we assume that its optimal value is known to us a priori. 
  \item Explore further branching heuristics for selecting free weights
\end{itemize}

To summarize, our proposed algorithm allows performing a greedy search based on consecutive results, which in turn helps to reduce the total solver computation time. Recap that the number of total combinations grows as $n^{\frac{n}{2}}$ where $n$ denotes the number of network weights. Hence, increasing the network size will eventually lead to an intractable amount of weight combinations to check. Therefore, we believe that utilizing a search tree and greedy methods for performing such a search is crucial.
\section{Conclusion and open questions}
\label{chap:conclusion}

\subsection{Conclusions}

In this research work, we demonstrate how off-the-shelf SMT solvers can be utilized to automatically repair unsafe NNs. We propose a few heuristics to preserve the similarity to the original networks and compare these heuristics. Then, to demonstrate its effectiveness, we compare our method with a naive baseline. Lastly, based on our observations, we propose a greedy search method for repairing the network (see \autoref{alg:first}). Applying our proposed repair method, ML practitioners should be capable of repairing small size unsafe neural networks. However, we expect our repair method to become intractable for larger networks unless other efficient strategies for traversing the search tree are explored, as suggested in \autoref{section:proposed-alg}. 

At first, our experiments show that the \textit{Samples} similarity heuristic outperforms other considered heuristics, in terms of both effectiveness and efficiency. Specifically, we observed how by freeing only a small subset of network weights (i.e., 1 or 2 weights), we are able to produce safe networks and even achieve substantial performance in terms of accuracy. Then, our second set of trials illustrates that our method is on par with a simple yet effective baseline, as measured by accuracy. Moreover, it also shows that there is enough space for improving the search procedure efficiency. For instance, by searching for a threshold value prior to the search process. Or, another possible heuristic is to traverse only certain parts of the search tree. Finally, we exploit these observations to later propose an algorithm sketch on \autoref{section:proposed-alg}. We also suggest a few strategies to further reduce the complexity of our proposed algorithm. Otherwise, the algorithm quickly becomes intractable when increasing the network size, or the number of free weights, as our analysis shows in this section. 

To summarize, in this work we define the task of repairing a neural network, we then propose a method to repair it. Later on, we demonstrate the capabilities of our method by performing extensive experiments, and empirically prove how it outperforms a simple baseline. To conclude, we provide an algorithm sketch (\autoref{alg:first}) for solving the general problem of repairing NNs with SMT solvers.

\subsection{Future work}

As follows from \autoref{section:proposed-alg}, improving the algorithm efficiency might be a good direction for future work, we proposed a few strategies that might achieve this. We expect these strategies to help with reducing the average exploration time of the proposed algorithm.  Then, it might be possible to explore more realistic benchmarks, such as \textit{Acas XU} networks we mentioned earlier. Once achieved, we may compare our framework to \cite{lin2020art}'s method which evolved from traditional gradient-based optimization, similarly to our baseline. Another possible trial is to compare their method to the current experiment results, mainly to find if it improves upon the naive baseline in terms of accuracy.

Additionally, in the spirit of related works such as \cite{sotoudeh2019correcting}, \cite{papusha2020incorrect}, the NN model complexity (in terms of its parameter number) can be further reduced using approximation methods. For example, one may use a smaller network (i.e., with fewer parameters) to approximate a larger one. We can also decompose the original NN into multiple, smaller NNs, that is, by partitioning its input or output spaces. For example, we may divide the input space into regions, then train multiple (less complex) models that correspond to each input partition. Another technique to reduce the output space which is common among ML practitioners is called one-vs.-all. Using this technique, we can train multiple binary classifiers instead of one multi-class classifier. 

A further dimension that can be possibly improved is the underlying SMT solver. As mentioned earlier, our theory solver mechanism is based on the \textit{CAD} algorithm. Perhaps the solver can be further tuned according to the trials results. Next, if such tools become more suitable for the task of synthesizing neural networks, then it might lead to the development of more efficient algorithms for producing provably correct NNs.  

Lastly, it may be possible to answer slightly different questions through our framework. For example, \textit{"can we find a new weights assignment such that the network accuracy becomes $100\%$?"}. Answering such questions might be useful for practitioners to decide if they need to employ a different NN architecture (for instance, one that has more capacity).
%
%

\begin{appendices}
\section{SMT Formula Encoding Example}
\label{appendix:basic_nn}

In this section, we describe a small neural network, a specification $\varphi$, and the corresponding SMT formula encoding to enforce that specification on a new neural network. 

\begin{figure}
	\centering
	\includegraphics[scale=0.55]{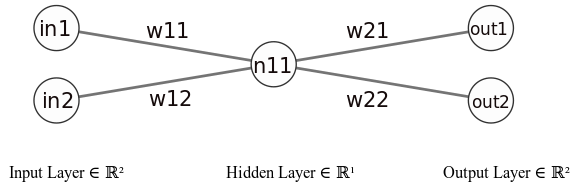}
	\caption{Two layer neural network with one ReLU neuron, and its corresponding labels describing the input nodes (with prefix \textbf{in}), ReLU neuron node (prefix \textbf{n}), output nodes (prefix \textbf{out}) and weights (prefix \textbf{w}). In addition there are three bias parameters (one for the ReLU neuron, and the other for the two output nodes) which are not depicted in this figure.}
	\label{fig:basicnn}
\end{figure}

Following our example in \autoref{chap:prelims}, we may use two-layer neural network with 1 ReLU neuron as depicted in Figure \ref{fig:basicnn}. The NN input is $x \in \mathbb{R}^2$, and its output is $y \in \mathbb{R}^2$. We shall call this network model A.

Let us assume model A does not meet the required specification $\varphi \equiv \; \forall in_1 \in [0, 0.5] \; . out_1 > out_2$. That is, a verification tool has found an assignment to the inputs $in_1, in_2$ in which $in_1 \in [0, 0.5]$ though $out_1 <= out_2$. We then search for a repaired NN - model B, by choosing to free two parameters, namely, $w_{21}$ and $bias_{21}$. In the following we describe the required encoding to formulate this search: 

\begin{equation*}
	\begin{aligned}
		\exists & {w_{21}, bias_{21}} \; ( \\ 
		  & \forall {in_1, in_2} \; ( & \\
		        & n_{11} = \text{if-then-else}( in_{1}*w_{11} + in_{2}*w_{12} + bias_{11} > 0, in_{1}*w_{11} + in_{2}*w_{12} + bias_{11}, 0) \; \land \\
		        & out_{1} = w_{21}*n_{11} + bias_{21} \; \land \\
		        & out_{2} = w_{22}*n_{11} + bias_{22} \; \land \\
		        & (in_{1} >= 0) \land (in_{1} <= 0.5) \implies (out_{1} > out_{2}) \\
		) \; )
	\end{aligned}
\end{equation*}

The variable $n_{11}$ denotes the ReLU value. The 'if-then-else' construct implements the selection of the two states that this neuron has. We include two parameters in our existential quantifier clause while the rest of them are assumed to be constant and equal to those of model A. Then, in case the solver returns SAT, it has found an assignment to $w_{21}, bias_{21}$ which satisfies the property $\varphi$. And in case it returns UNSAT, then there exists no assignment to $w_{21}, bias_{21}$ that satisfies $\varphi$. In this scenario, to repair the network, we may choose to free more weights, or other weights, until the solver returns SAT.

\end{appendices}

\bibliographystyle{unsrt}  
\bibliography{references}

\end{document}